\documentclass{article} % For LaTeX2e
\usepackage{iclr2017_conference,times}
\usepackage{hyperref}
\usepackage{url}
\usepackage[dvips]{graphicx}
\usepackage{amsmath}
\usepackage{amsthm}
\usepackage{amssymb}

\usepackage[small]{caption}

\title{Joint Multimodal Learning with Deep Generative Models}

\author{Masahiro Suzuki, Kotaro Nakayama, Yutaka Matsuo \\
  The University of Tokyo\\
  Bunkyo-ku, Tokyo, Japan \\
\texttt{\{masa,k-nakayama,matsuo\}@weblab.t.u-tokyo.ac.jp}
}

\begin{document}

\maketitle

\begin{abstract}
We investigate deep generative models that can exchange multiple modalities bi-directionally, e.g., generating images from corresponding texts and vice versa. Recently, some studies handle multiple modalities on deep generative models, such as variational autoencoders (VAEs). However, these models typically assume that modalities are forced to have a conditioned relation, i.e., we can only generate modalities in one direction. To achieve our objective, we should extract a joint representation that captures high-level concepts among all modalities and through which we can exchange them bi-directionally. As described herein, we propose a joint multimodal variational autoencoder (JMVAE), in which all modalities are independently conditioned on joint representation. In other words, it models a joint distribution of modalities. Furthermore, to be able to generate missing modalities from the remaining modalities properly, we develop an additional method, JMVAE-kl, that is trained by reducing the divergence between JMVAE's encoder and prepared networks of respective modalities. Our experiments show that our proposed method can obtain appropriate joint representation from multiple modalities and that it can generate and reconstruct them more properly than conventional VAEs. We further demonstrate that JMVAE can generate multiple modalities bi-directionally.

\end{abstract}

\vspace{-2mm}
\section{Introduction}
In our world, information is represented through various modalities. While images are represented by pixel information, these can also be described with text or tag information. People often exchange such information bi-directionally. For instance, we can not only imagine what ``a young female with a smile who does not wear glasses'' looks like, but also add this caption to a corresponding photograph. To do so, it is important to extract a joint representation that captures high-level concepts among all modalities. Then we can bi-directionally generate modalities through the joint representations. However, each modality typically has a different kind of dimension and structure, e.g., images (real-valued and dense) and texts (discrete and sparse). Therefore, the relations between each modality and the joint representations might become high nonlinearity. To discover such relations, deep neural network architectures have been used widely for multimodal learning \citep{Ngiam2011a, Srivastava2012}. The common approach with these models to learn joint representations is to share the top of hidden layers in modality specific networks. Among them, generative approaches using deep Boltzmann machines (DBMs) \citep{Srivastava2012, Sohn2014} offer the important advantage that these can generate modalities bi-directionally. 

Recently, variational autoencoders (VAEs) \citep{Welling2014,Rezende2014} have been proposed to estimate flexible deep generative models by variational inference methods. These models use back-propagation during training, so that it can be trained on large-scale and high-dimensional dataset compared with DBMs with MCMC training. Some studies have addressed to handle such large-scale and high-dimensional modalities on VAEs, but they are forced to model conditional distribution \citep{Kingma2014,Sohn2015,Pandey2016}. Therefore, it can only generate modalities in one direction. For example, we cannot obtain generated images from texts if we train the likelihood of texts given images. To generate modalities bi-directionally, all modalities should be treated equally under the learned joint representations, which is the same as previous multimodal learning models before VAEs.

As described in this paper, we develop a novel multimodal learning model with VAEs, which we call a joint multimodal variational autoencoder (JMVAE). The most significant feature of our model is that all modalities, $\mathbf{x}$ and $\mathbf{w}$ (e.g., images and texts), are conditioned independently on a latent variable $\mathbf{z}$ corresponding to joint representation, i.e., the JMVAE models a joint distribution of all modalities, $p(\mathbf{x},\mathbf{w})$. Therefore, we can extract a high-level representation that contains all information of modalities. Moreover, since it models a joint distribution, we can draw samples from both $p(\mathbf{x}|\mathbf{w})$ and $p(\mathbf{w}|\mathbf{x})$. Because, at this time, modalities that we want to generate are usually missing, the inferred latent variable becomes incomplete and generated samples might be collapsed in the testing time when missing modalities are high-dimensional and complicated. To prevent this issue, we propose a method of preparing the new encoders for each modality, $p(\mathbf{z}|\mathbf{x})$ and $p(\mathbf{z}|\mathbf{w})$, and reducing the divergence between the multimodal encoder  $p(\mathbf{z}|\mathbf{x},\mathbf{w})$, which we call JMVAE-kl. This contributes to more effective bi-directional generation of modalities, e.g., from face images to texts (attributes) and vice versa (see Figure 1).

The main contributions of this paper are as follows:
\begin{itemize}
\item We introduce a joint multimodal variational autoencoder (JMVAE), which is the first study to train joint distribution of modalities with VAEs.
\item We propose an additional method (JMVAE-kl), which prevents generated samples from being collapsed when some modalities are missing. We experimentally confirm that this method solves this issue.
\item We show qualitatively and quantitatively that JMVAE can extract appropriate joint distribution and that it can generate and reconstruct modalities similarly or more properly than conventional VAEs. 
\item We demonstrate that the JMVAE can generate multiple modalities bi-directionally even if these modalities have completely different kinds of dimensions and structures, e.g., high-dimentional color face images and low-dimentional binary attributes.
\end{itemize}

\begin{figure}[t]
 \begin{center}
  \includegraphics[scale=0.68]{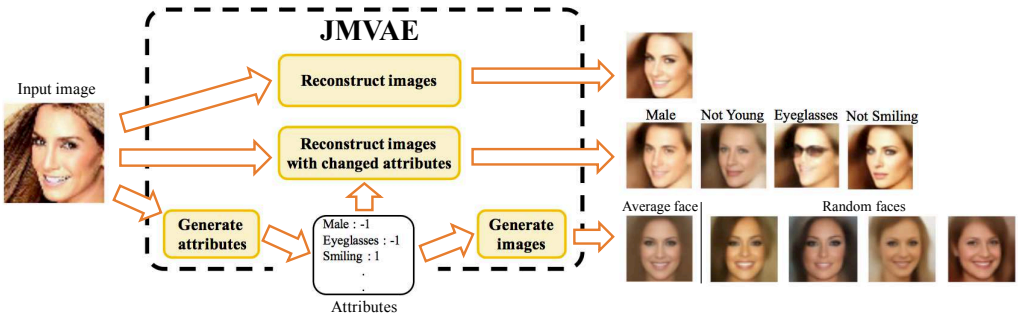}
 \end{center}
 \caption{Various images and attributes generated from an input image. We used the CelebA dataset \citep{Eyebrows2015} to train and test models in this example. Each yellow box corresponds to different processes. All processes are estimated from a single generative model: the joint multimodal variational autoencoder (JMVAE), which is our proposed model. }
 \label{fig:overview} 
\end{figure}
\vspace{-2mm}
\section{Related work}
The common approach of multimodal learning with deep neural networks is to share the top of hidden layers in modality specific networks. \citet{Ngiam2011a} proposed this approach with deep autoencoders (AEs) and found that it can extract better representations than single modality settings. \citet{Srivastava2012} also took this idea but used deep Boltzmann machines (DBMs) \citep{salakhutdinov2009deep}. DBMs are generative models with undirected connections based on maximum joint likelihood learning of all modalities. Therefore, this model can generate modalities bi-directionally. \citet{Sohn2014} improved this model to exchange multiple modalities effectively, which are based on minimizing the variation of information and JMVAE-kl in ours can be regarded as minimizing it by variational inference methods (see Section \ref{sec:inf_rec} and Appendix \ref{sec:derivation}). However, it is computationally difficult for DBMs to train high-dimensional data such as natural images because of MCMC training.

Recently, VAEs \citep{Welling2014,Rezende2014} are used to train such high-dimensional modalities. \citet{Kingma2014,Sohn2015} propose conditional VAEs (CVAEs), which maximize a conditional log-likelihood by variational methods. Many studies are based on CVAEs to train various multiple modalities such as handwriting digits and labels \citep{Kingma2014,Sohn2015}, object images and degrees of rotation \citep{kulkarni2015deep}, face images and attributes \citep{Larsen2015, yan2015attribute2image}, and natural images and captions \citep{mansimov2015generating}. The main features of CVAEs are that the relation between modalities is one-way and a latent variable does not contain the information of a conditioned modality\footnote{According to \citet{Louizos2015}, this independence might not be satisfied strictly because the encoder in CVAEs still has the dependence.}, which are unsuitable for our objective.

\citet{Pandey2016} proposed a conditional multimodal autoencoder (CMMA), which also maximizes the conditional log-likelihood. The difference between CVAEs is that a latent variable is connected directly from a conditional variable, i.e., these variables are not independent. Moreover, this model forces the latent representation from an input to be close to the joint representation from multiple inputs, which is similar to JMVAE-kl. However, the CMMA still considers that modalities are generated in fixed direction. This is the most different part from ours.
\vspace{-2mm}
\section{Methods}
This section first introduces the algorithm of VAEs briefly and then proposes a novel multimodal learning model with VAEs, which we call the joint multimodal variational autoencoder (JMVAE).

\vspace{-2mm}
\subsection{Variational autoencoders}
Given observation variables $\mathbf{x}$ and corresponding latent variables $\mathbf{z}$, their generating processes are definable as $\mathbf{z}\sim p(\mathbf{z})=\mathcal{N}(\mathbf{0},\mathbf{I})$ and $\mathbf{x}\sim p_\theta(\mathbf{x}|\mathbf{z})$, where $\theta$ is the model parameter of $p$. The objective of VAEs is maximization of the marginal distribution $p(\mathbf{x})=\int p_\theta(\mathbf{x}|\mathbf{z})p(\mathbf{z})d\mathbf{x}$. Because this distribution is intractable, we instead train the model to maximize the following lower bound of the marginal distribution $\mathcal{L}_{VAE}(\mathbf{x})$ as
\begin{eqnarray}
\log p(\mathbf{x})\geq  -D_{KL}(q_\phi(\mathbf{z}|\mathbf{x})||p(\mathbf{z})) + E_{q_\phi(\mathbf{z}|\mathbf{x})}[\log p_\theta(\mathbf{x}|\mathbf{z})] = \mathcal{L}_{VAE}(\mathbf{x}),
\label{eq:VAE_lower}
\end{eqnarray}
where $q_\phi(\mathbf{z}|\mathbf{x})$ is an approximate distribution of posterior $p(\mathbf{z}|\mathbf{x})$ and $\phi$ is the model parameter of $q$. We designate $q_\phi(\mathbf{z}|\mathbf{x})$ as {\rm encoder} and $p_\theta(\mathbf{x}|\mathbf{z})$ as {\rm decoder}. Moreover, in Equation \ref{eq:VAE_lower}, the first term represents a regularization. The second one represents a negative reconstruction error.

To optimize the lower bound $\mathcal{L}(\mathbf{x})$ with respect to parameters $\theta,\phi$, we estimate gradients of Equation \ref{eq:VAE_lower} using stochastic gradient variational Bayes (SGVB). If we consider $q_\phi(\mathbf{z}|\mathbf{x})$ as Gaussian distribution $\mathcal{N}(\mathbf{z};{\boldsymbol \mu},{\rm diag} ({\boldsymbol \sigma}^2))$, where $\phi=\{{\boldsymbol \mu},{\boldsymbol \sigma}^2\}$, then we can reparameterize $\mathbf{z}\sim q_\phi(\mathbf{z}|\mathbf{x})$ to $\mathbf{z}={\boldsymbol \mu}+{\boldsymbol \sigma} \odot{\boldsymbol \epsilon}$, where ${\boldsymbol \epsilon}\sim \mathcal{N}(\mathbf{0},\mathbf{I})$. Therefore, we can estimate the gradients of the negative reconstruction term in Equation \ref{eq:VAE_lower} with respect to $\theta$ and $\phi$ as $\nabla_{\theta, \phi} E_{q_\phi(\mathbf{z}|\mathbf{x})}[\log p_\theta(\mathbf{x}|\mathbf{z})] = E_{\mathcal{N}({\boldsymbol \epsilon};\mathbf{0},\mathbf{I})}[\nabla_{\theta, \phi}\log p_\theta(\mathbf{z}|{\boldsymbol \mu}+{\boldsymbol \sigma}\odot {\boldsymbol \epsilon})]$. Because the gradients of the regularization term are solvable analytically, we can optimize Equation \ref{eq:VAE_lower} with standard stochastic optimization methods.

\vspace{-2mm}
\subsection{Joint Multimodal variational autoencoders}
\begin{figure}[tb]
 		\begin{center}
		\includegraphics[scale=0.61]{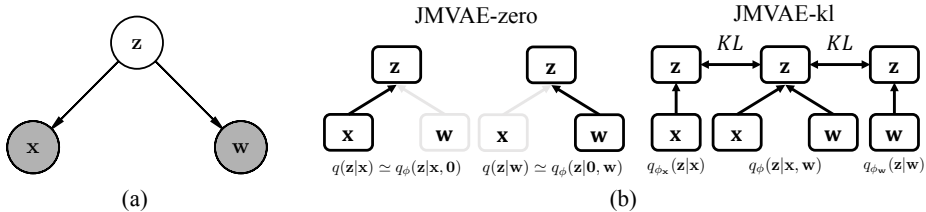}
		\end{center}
		\caption{(a) Graphical model of the JMVAE. Gray circles represent observed variables. The white one denotes a latent variable. (b) Two approaches to estimate encoders with a single input, $q(\mathbf{z}|\mathbf{x})$ and $q(\mathbf{z}|\mathbf{w})$, on the JMVAE: {\sl left}, make modalities except an input modality missing ({\sl JMVAE-zero}); {\sl right}, prepare encoders that have a single input and make them close to the JMVAE encoder ({\sl JMVAE-kl}).}
		\label{fig:approaches}
\end{figure}

Next, we consider $i.i.d.$ dataset $(\mathbf{X}, \mathbf{W})=\{(\mathbf{x}_1,\mathbf{w}_1),...,(\mathbf{x}_N,\mathbf{w}_N)\}$, where two modalities $\mathbf{x}$ and $\mathbf{w}$ have different kinds of dimensions and structures\footnote{In our experiment, these depend on dataset, see Section \ref{sec:model_arc}.}. Our objective is to generate two modalities bi-directionally. For that reason, we assume that these are conditioned independently on the same latent concept $\mathbf{z}$: joint representation. Therefore, we assume their generating processes as $\mathbf{z} \sim p(\mathbf{z})$ and $\mathbf{x},\mathbf{w} \sim p(\mathbf{x},\mathbf{w}|\mathbf{z})=p_{\theta_\mathbf{x}}(\mathbf{x}|\mathbf{z})p_{\theta_\mathbf{w}}(\mathbf{w}|\mathbf{z})$, where $\theta_\mathbf{x}$ and $\theta_\mathbf{w}$ represent the model parameters of each independent $p$. Figure \ref{fig:approaches}(a) shows a graphical model that represents generative processes. One can see that this models joint distribution of all modalities, $p(\mathbf{x},\mathbf{w})$. Therefore, we designate this model as a {\sl joint multimodal variational autoencoder} (JMVAE).

Considering an approximate posterior distribution as $q_\phi(\mathbf{z}|\mathbf{x},\mathbf{w})$, we can estimate a lower bound of the log-likelihood $\log p(\mathbf{x},\mathbf{w})$ as follows:
\begin{eqnarray}
  \mathcal{L}_{JM}(\mathbf{x},\mathbf{w}) &=& E_{q_\phi(\mathbf{z}|\mathbf{x},\mathbf{w})}[\log\frac{p_\theta(\mathbf{x},\mathbf{w},\mathbf{z})}{q_\phi(\mathbf{z}|\mathbf{x},\mathbf{w})}] \\
\label{eq:JMVAE_log_likelihood}
&=& -D_{KL}(q_\phi(\mathbf{z}|\mathbf{x},\mathbf{w})||p(\mathbf{z})) \nonumber \\ 
& & + E_{q_\phi(\mathbf{z}|\mathbf{x},\mathbf{w})}[\log p_{\theta_\mathbf{x}}(\mathbf{x}|\mathbf{z})] + E_{q_\phi(\mathbf{z}|\mathbf{x},\mathbf{w})}[\log p_{\theta_\mathbf{w}}(\mathbf{w}|\mathbf{z})].
\label{eq:JMVAE_lower}
\end{eqnarray}
Equation \ref{eq:JMVAE_lower} has two negative reconstruction terms which are correspondent to each modality. As with VAEs, we designate $q_\phi(\mathbf{z}|\mathbf{x},\mathbf{w})$ as the encoder and both $p_{\theta_\mathbf{x}}(\mathbf{x}|\mathbf{z})$ and $p_{\theta_\mathbf{w}}(\mathbf{w}|\mathbf{z})$ as decoders. 

We can apply the SGVB to Equation \ref{eq:JMVAE_lower} just as Equation \ref{eq:VAE_lower}, so that we can parameterize the encoder and decoder as deterministic deep neural networks and optimize them with respect to their parameters, $\theta_\mathbf{x}$, $\theta_\mathbf{w}$, and $\phi$. Because each modality has different feature representation, we should set different networks for each decoder, $p_{\theta_\mathbf{x}}(\mathbf{x}|\mathbf{z})$ and $p_{\theta_\mathbf{w}}(\mathbf{w}|\mathbf{z})$. The type of distribution and corresponding network architecture depends on the representation of each modality, e.g., Gaussian when the representation of modality is continuous, and a Bernoulli when it is a binary value.

Unlike original VAEs and CVAEs, the JMVAE models joint distribution of all modalities. In this model, modalities are conditioned independently on a joint latent variable. Therefore, we can extract better representation that includes all information of modalities. Moreover, we can estimate both marginal distribution and conditional distribution in bi-directional, so that we can not only obtain images reconstructed themselves but also draw texts from corresponding images and vice versa. Additionally, we can extend JMVAEs to handle more than two modalities such as $p(\mathbf{x},\mathbf{w}_1,\mathbf{w}_2,...)$ in the same learning framework.

\vspace{-2mm}
\subsection{Inference missing modalities}
\vspace{-2mm}
\label{sec:inf_rec}
In the JMVAE, we can extract joint latent features by sampling from the encoder $q_{\phi}(\mathbf{z}|\mathbf{x},\mathbf{w})$ at testing time. Our objective is to exchange modalities bi-directionally, e.g., images to texts and vice versa. In this setting, modalities that we want to sample are missing, so that inputs of such modalities are set to zero (the left panel of Figure \ref{fig:approaches}(b)). The same is true of reconstructing a modality only from itself. This is a natural way in discriminative multimodal settings to estimate samples from unimodal information \citep{Ngiam2011a}. However, if missing modalities are high-dimensional and complicated such as natural images, then the inferred latent variable becomes incomplete and generated samples might collapse. 

We propose a method to solve this issue, which we designate as JMVAE-kl. Moreover, we describe the former way as JMVAE-zero to distinguish it. Suppose that we have encoders with a single input, $q_{\phi_\mathbf{x}}(\mathbf{z}|\mathbf{x})$ and $q_{\phi_\mathbf{w}}(\mathbf{z}|\mathbf{w})$, where $\phi_\mathbf{x}$ and $\phi_\mathbf{w}$ are parameters. We would like to train them by bringing their encoders close to an encoder $q_{\phi}(\mathbf{z}|\mathbf{x},\mathbf{w})$ (the right panel of Figure \ref{fig:approaches}(b)). Therefore, the object function of JMVAE-kl becomes
\begin{eqnarray}
  \mathcal{L}_{JM_{kl}(\alpha)}(\mathbf{x},\mathbf{w}) = \mathcal{L}_{JM}(\mathbf{x},\mathbf{w}) - \alpha \cdot [D_{KL}(q_\phi(\mathbf{z}|\mathbf{x},\mathbf{w})||q_{\phi_\mathbf{x}}(\mathbf{z}|\mathbf{x}))+D_{KL}(q_\phi(\mathbf{z}|\mathbf{x},\mathbf{w})||q_{\phi_\mathbf{w}}(\mathbf{z}|\mathbf{w}))],
\label{eq:new_lowerbound}
\end{eqnarray}
where $\alpha$ is a factor that regulates the KL divergence terms.

From another viewpoint, maximizing Equation \ref{eq:new_lowerbound} can be regarded as minimizing the variation of information (VI) by variational inference methods (proven and derived in Appendix \ref{sec:derivation}). The VI, a measure of the distance between two variables, is written as $-E_{p_D(\mathbf{x},\mathbf{w})}[\log p(\mathbf{x}|\mathbf{w})+\log p(\mathbf{w}|\mathbf{x})]$, where $p_D$ is the data distribution. It is apparent that the VI is the sum of two negative conditional log-likelihoods. Therefore, minimizing the VI contributes to appropriate bi-directional exchange of modalities. \citet{Sohn2014} also train their model to minimize the VI for the same objective as ours. However, they use DBMs with MCMC training.

\section{Experiments}
This section presents evaluation of the qualitative and quantitative performance and confirms the JMVAE functionality in practice.
\vspace{-2mm}
\subsection{Datasets}
As described herein, we used two datasets: MNIST and CelebA \citep{Eyebrows2015}.

MNIST is not a dataset for multimodal setting. In this work, we used this dataset for toy problem of multimodal learning. We consider handwriting images and corresponding digit labels as two different modalities. We used 50,000 as training set and the remaining 10,000 as a test set.

CelebA consists of 202,599 color face images and corresponding 40 binary attributes such as male, eyeglasses,  and mustache. In this work, we regard them as two modalities. This dataset is challenging because these have completely different kinds of dimensions and structures. Beforehand, we cropped the images to squares and resized to 64 $\times$ 64 and normalized. From the dataset, we chose 191,899 images that are identifiable face by OpenCV and used them for our experiment. We used 90\% out of all the dataset contains as training set and the remaining 10\% of them as test set.
\vspace{-2mm}
\subsection{Model architectures}
\label{sec:model_arc}

For MNIST, we considered images as $\mathbf{x}\in \mathbb{R}^{28\times 28}$ and corresponding labels as $\mathbf{w}\in \{0,1\}^{10}$. We prepared two networks each with two dense layers of 512 hidden units and using leaky rectifiers and shared the top of each layers and mapped them into 64 hidden units. Moreover, we prepared two networks each with three dense layers of 512 units and set $p(\mathbf{x}|\mathbf{z})$ as Bernoulli and $p(\mathbf{w}|\mathbf{z})$ as categorical distribution whose output layer is softmax. We used warm-up \citep{bowman2015generating, Sonderby2016a}, which first forces training only of the term of the negative reconstruction error and then gradually increases the effect of the regularization term to prevent local minima during early training. We increased this term linearly during the first $N_t$ epochs as with \citet{Sonderby2016a}. We set $N_t=200$ and trained for $500$ epochs on MNIST. Moreover, same as \citet{Burda2015,Sonderby2016a}, we resampled the binarized training values randomly from MNIST for each epoch to prevent over-fitting.

For CelebA, we considered face images as $\mathbf{x}\in \mathbb{R}^{32\times 32\times 3}$ and corresponding attributes as $\mathbf{w}\in \{-1,1\}^{40}$. We prepared two networks with layers (four convolutional and a flattened layers for $\mathbf{x}$ and two dense layers for $\mathbf{w}$) with ReLU and shared the top of each layers and mapped them into 128 units. For the decoder, we prepared two networks, with a dense and four deconvolutional layers for $\mathbf{x}$ and three dense layers for $\mathbf{w}$, and set Gaussian distribution for decoder of both modalities, where the variance of Gaussian was fixed to 1 for the decoder of $\mathbf{w}$. In CelebA settings, we combined JMVAE with generative adversarial networks (GANs) \citep{Goodfellow2014} to generate clearer images. We considered the network of $p(\mathbf{x}|\mathbf{z})$ as {\rm generator} in GAN, then we optimized the GAN loss with the lower bound of the JMVAE, which is the same way as a VAE-GAN model \citep{Larsen2015}. As presented herein, we describe this model as JMVAE-GAN. We set $N_t=20$ and trained for $100$ epochs on CelebA.

We used the Adam optimization algorithm \citep{Kingma2015a} with a learning rate of $10^{-3}$ on MNIST and $10^{-4}$ on CelebA. The models were implemented using Theano \citep{TheTheanoDevelopmentTeam2016} and Lasagne \citep{Dieleman2015}.% and Tars\footnote{https://github.com/masa-su/Tars}.

\vspace{-2mm}
\subsection{Quantitative evaluation}
\label{sec:quan}
\subsubsection{Evaluation method}
\label{sec:eval}
For this experiment, we estimated test log-likelihood to evaluate the performance of model. This estimate roughly corresponds to negative reconstruction error. Therefore, higher is better. From this performance, we can find that not only whether the JMVAE can generate samples properly but also whether it can obtain joint representation properly. If the log-likelihood of a modality is low, representation for this modality might be hurt by other modalities. By contrast, if it is the same or higher than model trained on a single modality, then other modalities contribute to obtaining appropriate representation.

We estimate the test marginal log-likelihood and test conditional log-likelihood on JMVAE. We compare the test marginal log-likelihood against VAEs \citep{Welling2014,Rezende2014} and the test conditional log-likelihood against CVAEs \citep{Kingma2014,Sohn2015} and CMMAs \citep{Pandey2016}. On CelebA, we combine all competitive models with GAN and describe them as VAE-GAN, CVAE-GAN, and CMMA-GAN. For fairness, architectures and parameters of these competitive models were set to be as close as possible to those of JMVAE.

We calculate the importance weighted estimator \citep{Burda2015} from lower bounds at testing time because we would like to estimate the true test log-likelihood from lower bounds. To estimate the test marginal log-likelihood $p(\mathbf{w})$ of the JMVAE, we use two possible lower bounds: sampling from $q_{\phi}(\mathbf{z}|\mathbf{x},\mathbf{w})$ or $q_{\phi_x}(\mathbf{z}|\mathbf{x})$. We describe the former lower bound as the {\sl multiple} lower bound and the latter one as the {\sl single} lower bound. When we estimate the test conditional log-likelihood $\log p(\mathbf{x}|\mathbf{w})$, we also use two lower bounds, each of which is estimated by sampling from $q_{\phi}(\mathbf{z}|\mathbf{x},\mathbf{w})$ (multiple) or $q_{\phi_w}(\mathbf{z}|\mathbf{w})$ (single) (see Appendix \ref{sec:app1} for more details). To estimate the single lower bound, we should approximate the single encoder ($q_{\phi_x}(\mathbf{z}|\mathbf{x})$ or $q_{\phi_w}(\mathbf{z}|\mathbf{w})$) by JMVAE-zero or JMVAE-kl. When the value of log-likelihood with the single lower bound is the same or larger than that with the multiple lower bound, the approximation of the single encoder is good. Note that original VAEs use a single lower bound and that CVAEs and CMMAs use a multiple lower bound.

\vspace{-2mm}
\subsubsection{MNIST}

\begin{table}[tb]
    \caption{Evaluation of test log-likelihood. All models are trained and tested on MNIST. $\alpha$ is a coefficient of regularization term in JMVAE-kl (Equation \ref{eq:new_lowerbound}): {\sl left}, marginal log-likelihood; {\sl right}, conditional log-likelihood.}
  \begin{center}
  \small
  \begin{tabular}{lcc}\hline
     & \multicolumn{2}{c}{$\leq\log p(\mathbf{x})$} \\\cline{2-3}
     & multiple & single\\\hline
     VAE & & -86.91 \\\hline
     JMVAE-zero & -86.89 & -86.89 \\
	 JMVAE-kl, $\alpha=0.01$ & -86.89 & {\bf -86.55} \\
     JMVAE-kl, $\alpha=0.1$ & -86.86 & -86.73 \\
     JMVAE-kl, $\alpha=1$ & -89.20 & -89.20 \\\hline
  \end{tabular}
  \begin{tabular}{lcc}\hline
     & \multicolumn{2}{c}{$\leq\log p(\mathbf{x}|\mathbf{w})$} \\\cline{2-3}
     &  multiple & single\\\hline
     CVAE & {\bf -83.80} & \\
     CMMA & -86.12 & \\\hline
     JMVAE-zero & -84.64 & -4838 \\
     JMVAE-kl, $\alpha=0.01$ & -84.61 & -129.6 \\
     JMVAE-kl, $\alpha=0.1$ & -84.72 & -126.0 \\
     JMVAE-kl, $\alpha=1$ & -86.97 & -112.7  \\\hline
  \end{tabular}
  \label{tab:mnist_log-likelihood}
  \end{center}
\end{table}

\begin{table}[tb]
   \caption{Evaluation of log-likelihood. Models are trained and tested on CelebA. We trained JMVAE-kl and set $\alpha=0.1$: {\sl left}, marginal log-likelihood; {\sl right}, conditional log-likelihood (with the multiple lower bound).}
  \begin{center}
  \small
  \begin{tabular}{lcc}\hline
      & \multicolumn{2}{c}{$\leq\log p(\mathbf{x})$} \\\cline{2-3}
      & multiple & single\\\hline
     VAE-GAN & & -4439 \\\hline
     JMVAE-GAN & {\bf -4141} & -4144  \\\hline
  \end{tabular}
  \begin{tabular}{lc}\hline
     & $\leq\log p(\mathbf{x}|\mathbf{w})$\\ \hline
    CVAE-GAN & -4152 \\ 
    CMMA-GAN & -4147\\\hline
    JMVAE-GAN & {\bf -4130}\\\hline
  \end{tabular}
  \end{center}
  \label{tab:celeba_log-likelihood}
\end{table}

Our first experiment evaluated the test marginal log-likelihood and compared it with that of the VAE on MNIST dataset. We trained the model with both JMVAE-zero and JMVAE-kl and confirmed these differences. As described in Section \ref{sec:eval}, we have two possible ways of estimating the marginal log-likelihood of the JMVAE, i.e., multiple and single lower bounds. The left of Table \ref{tab:mnist_log-likelihood} shows the test marginal log-likelihoods of the VAE and JMVAE. It is apparent that log-likelihood of the JMVAE-zero is the same or slightly better than that of the VAE. In the case of the log-likelihood of JMVAE-kl, the log-likelihood becomes better as $\alpha$ is small. Especially, JMVAE-kl with $\alpha=0.01$ and single lower bound archives the highest log-likelihood in Table \ref{tab:mnist_log-likelihood}. If $\alpha$ is 1, however, then the test log-likelihood on JMVAE-kl becomes much lower. This is because the influence of the regularization term becomes strong as $\alpha$ is large. 

Next, we evaluated the test conditional log-likelihood and compared it with that of the CVAE and CMMA conditioned on $\mathbf{w}$. As in the case of the marginal log-likelihood, we can estimate the JMVAE's conditional log-likelihood by both the single and multiple lower bound. The single bound can be estimated using JMVAE-zero or JMVAE-kl. The right of Table \ref{tab:mnist_log-likelihood} shows the test conditional log-likelihoods of the JMVAE, CVAE, and CMMA. It is apparent that the CVAE achieves the highest log-likelihood. Even so, in the case of multiple bound, log-likelihoods with both JMVAE-zero and JMVAE-kl (except $\alpha=1$) outperform that of the CMMA. 

It should be noted that the log-likelihood with JMVAE-zero and single bound is significantly low. As described in Section \ref{sec:inf_rec}, this is because a modality $\mathbf{w}$ is missing as input. By contrast, it is apparent that the log-likelihood with JMVAE-kl is improved significantly from that with JMVAE-zero. It shows that JMVAE-kl solves the issue of missing modalities (we can also find this result in generated images, see Appendix \ref{sec:mnist_cond}). Moreover, we find that this log-likelihood becomes better as $\alpha$ is large, which is opposite to the other results. Therefore, there is a trade-off between whether each modality can be reconstructed properly and whether multiple modalities can be exchanged properly and it can be regulated by $\alpha$.

\vspace{-2mm}
\subsubsection{CelebA}
In this section, we used CelebA dataset to evaluate the JMVAE. Table \ref{tab:celeba_log-likelihood} presents the evaluations of marginal and conditional log-likelihood. From this table, it is apparent that values of both marginal and conditional log-likelihood with JMVAEs are larger than those with other competitive methods. Moreover, comparison with Table \ref{tab:mnist_log-likelihood} shows that the improvement on CelebA is greater than that on MNIST, which suggests that joint representation with multiple modalities contributes to improvement of the quality of the reconstruction and generation in the case in which an input modality is large-dimensioned and complicated.

\vspace{-2mm}
\subsection{Qualitative evaluation}
\vspace{-1mm}
\subsubsection{Joint representation on MNIST}
\begin{figure}[tb]
 \begin{center}
  \includegraphics[scale=0.69]{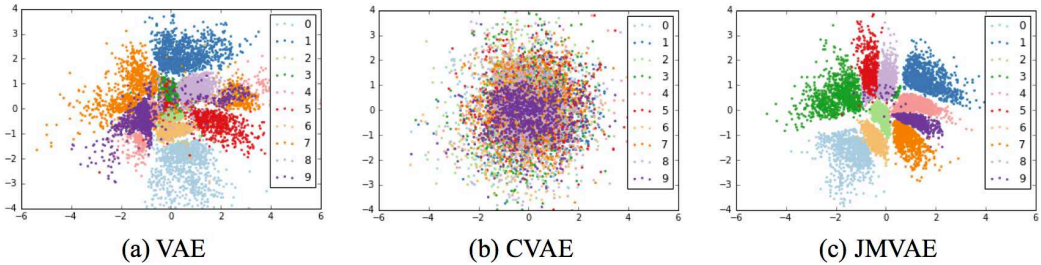}
 \end{center}
 \caption{Visualizations of 2-D latent representation. The network architectures are the same as those in Section \ref{sec:quan}, except that the dimension of the top hidden layer is forced into 2. Points with different colors correspond to the digit labels. These were sampled from $q(\mathbf{z}|\mathbf{x})$ in the VAE and $q(\mathbf{z}|\mathbf{x}, \mathbf{w})$ in both the CVAE and JMVAE. We used JMVAE-zero as the JMVAE.}
 \label{fig:px_multimodal}
 \end{figure}
 
In this section, we first evaluated that the JMVAE can obtain joint representation that includes the information of modalities. Figure \ref{fig:px_multimodal} shows the visualization of latent representation with the VAE, CVAE, and JMVAE on MNIST. It is apparent that the JMVAE obtains more discriminable latent representation by adding digit label information. Figure \ref{fig:px_multimodal}(b) shows that, in spite of using multimodal information as with the JMVAE, points in CVAE are distributed irrespective of labels because CVAEs force latent representation to be independent of label information, i.e., it is not objective for CVAEs to obtain joint representation.

\vspace{-2mm}
\subsubsection{Generating faces from attributes and joint representation on CelebA}
\begin{figure}[t]
 \begin{center}
  \includegraphics[scale=0.68]{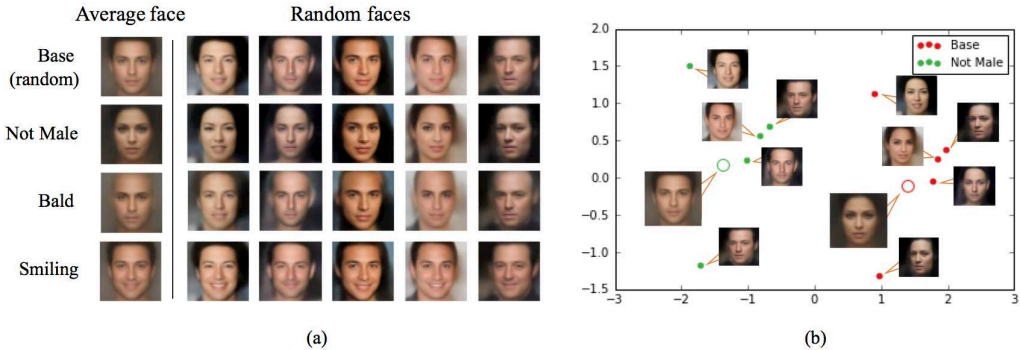}
 \end{center}
 \caption{(a) Generation of average faces and corresponding random faces. We first set all values of attributes $\{-1, 1\}$ randomly and designate them as Base. Then, we choose an attribute that we want to set (e.g., Male, Bald, Smiling) and change this value in Base to $2$ (or $-2$ if we want to set "Not"). Each column corresponds to same attribute according to legend. Average faces are generated from $p(\mathbf{x}|\mathbf{z}_{mean})$, where $\mathbf{z}_{mean}$ is a mean of $q(\mathbf{z}|\mathbf{w})$. Moreover, we can obtain various images conditioned on the same values of attributes such as $\mathbf{x}\sim p(\mathbf{x}|\mathbf{z})$, where $\mathbf{z}=\mathbf{z}_{mean}+{\boldsymbol \sigma}\odot{\boldsymbol \epsilon}$, ${\boldsymbol \epsilon}\sim \mathcal{N}(\mathbf{0},{\boldsymbol \zeta})$, and ${\boldsymbol \zeta}$ is the parameter which determines the range of variance. In this figure, we set ${\boldsymbol \zeta}=0.6$. Each row in random faces has the same ${\boldsymbol \epsilon}$. (b) PCA visualizations of latent representation. Colors indicate which attribute each sample is conditioned on.}
 \label{fig:mean_celeba}
\end{figure}

Next, we confirm that JMVAE-GAN on CelebA can generate images from attributes. Figure \ref{fig:mean_celeba}(a) portrays generated faces conditioned on various attributes. We find that we can generate an average face of each attribute and various random faces conditioned on a certain attributes. Figure \ref{fig:mean_celeba}(b) shows that samples are gathered for each attribute and that locations of each variation are the same irrespective of attributes. From these results, we find that manifold learning of joint representation with images and attributes works well.

\vspace{-2mm}
\subsubsection{Bi-directional generation between faces and attributes on CelebA}
\begin{figure}[t]
 \begin{center}
  \includegraphics[scale=0.68]{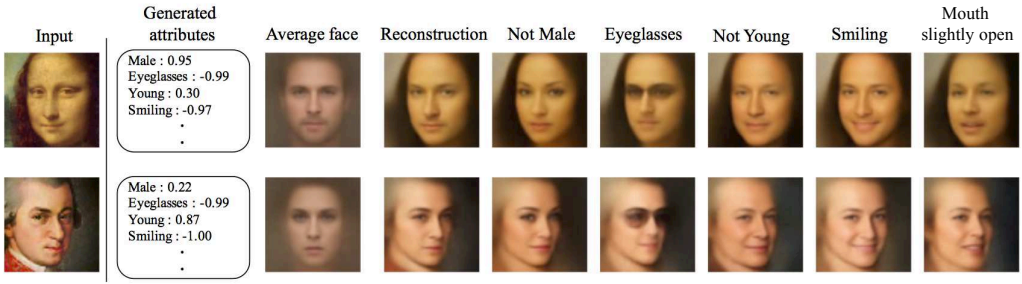}
 \end{center}
 \caption{Portraits of the Mona Lisa\protect\footnotemark (upper) and Mozart\protect\footnotemark  (lower), generated their attributes, and reconstructed images conditioned on varied attributes, according to the legend. We cropped and resized it in the same way as CelebA. The procedure is as follows: generate the corresponding attributes $\mathbf{w}$ from an unlabeled image $\mathbf{x}$; generate an average face $\mathbf{x}_{mean}$ from the attributes $\mathbf{w}$; select attributes which we want to vary and change the values of these attributes; generate the changed average face $\mathbf{x}'_{mean}$ from the changed attributes; and obtain a changed reconstruction image $\mathbf{x}'$ by $\mathbf{x}+\mathbf{x}'_{mean}-\mathbf{x}_{mean}$.}
 \label{fig:change_monalisa_celeba}
\end{figure}
\footnotetext{https://en.wikipedia.org/wiki/Mona\_Lisa}
\footnotetext{https://en.wikipedia.org/wiki/Wolfgang\_Amadeus\_Mozart}
Finally, we demonstrate that JMVAE-GAN can generate bi-directionally between faces and attributes. Figure \ref{fig:change_monalisa_celeba} shows that MVAE-GAN can generate both attributes and changed images conditioned on various attributes from images which had no attribute information. This way of generating an image by varying attributes is similar to the way of the CMMA \citep{Pandey2016}. However, the CMMA cannot generate attributes from an image because it only generates images from attributes in one direction.

\section{Conclusion and Future Work}
In this paper, we introduced a novel multimodal learning model with VAEs, the joint multimodal variational autoencoders (JMVAE). In this model, modalities are conditioned independently on joint representation, i.e., it models a joint distribution of all modalities. We further proposed the method (JMVAE-kl) of reducing the divergence between JMVAE's encoder and a prepared encoder of each modality to prevent generated samples from collapsing when modalities are missing. We confirmed that the JMVAE can obtain appropriate joint representations and high log-likelihoods on MNIST and CelebA datasets. Moreover, we demonstrated that the JMVAE can generate multiple modalities bi-directionally on the CelebA dataset.

In future work, we would like to evaluate the multimodal learning performance of JMVAEs using various multimodal datasets such as containing three or more modalities.

%\subsubsection*{Acknowledgments}

\bibliography{iclr2017_conference.bib}

\begin{thebibliography}{22}
\providecommand{\natexlab}[1]{#1}
\providecommand{\url}[1]{\texttt{#1}}
\expandafter\ifx\csname urlstyle\endcsname\relax
  \providecommand{\doi}[1]{doi: #1}\else
  \providecommand{\doi}{doi: \begingroup \urlstyle{rm}\Url}\fi

\bibitem[Bowman et~al.(2015)Bowman, Vilnis, Vinyals, Dai, Jozefowicz, and
  Bengio]{bowman2015generating}
Samuel~R Bowman, Luke Vilnis, Oriol Vinyals, Andrew~M Dai, Rafal Jozefowicz,
  and Samy Bengio.
\newblock Generating sentences from a continuous space.
\newblock \emph{arXiv preprint arXiv:1511.06349}, 2015.

\bibitem[Burda et~al.(2015)Burda, Grosse, and Salakhutdinov]{Burda2015}
Yuri Burda, Roger Grosse, and Ruslan Salakhutdinov.
\newblock Importance weighted autoencoders.
\newblock \emph{arXiv preprint arXiv:1509.00519}, 2015.

\bibitem[Dieleman et~al.(2015)Dieleman, Schlüter, Raffel, Olson, Sønderby,
  Nouri, Maturana, Thoma, Battenberg, Kelly, Fauw, Heilman, de~Almeida, McFee,
  Weideman, Takács, de~Rivaz, Crall, Sanders, Rasul, Liu, French, and
  Degrave]{Dieleman2015}
Sander Dieleman, Jan Schlüter, Colin Raffel, Eben Olson, Søren~Kaae
  Sønderby, Daniel Nouri, Daniel Maturana, Martin Thoma, Eric Battenberg, Jack
  Kelly, Jeffrey~De Fauw, Michael Heilman, Diogo~Moitinho de~Almeida, Brian
  McFee, Hendrik Weideman, Gábor Takács, Peter de~Rivaz, Jon Crall, Gregory
  Sanders, Kashif Rasul, Cong Liu, Geoffrey French, and Jonas Degrave.
\newblock Lasagne: First release., August 2015.
\newblock URL \url{http://dx.doi.org/10.5281/zenodo.27878}.

\bibitem[Goodfellow et~al.(2014)Goodfellow, Pouget-Abadie, Mirza, Xu,
  Warde-Farley, Ozair, Courville, and Bengio]{Goodfellow2014}
Ian Goodfellow, Jean Pouget-Abadie, Mehdi Mirza, Bing Xu, David Warde-Farley,
  Sherjil Ozair, Aaron Courville, and Yoshua Bengio.
\newblock Generative adversarial nets.
\newblock In \emph{Advances in Neural Information Processing Systems}, pp.\
  2672--2680, 2014.

\bibitem[Kingma \& Ba(2014)Kingma and Ba]{Kingma2015a}
Diederik Kingma and Jimmy Ba.
\newblock Adam: A method for stochastic optimization.
\newblock \emph{arXiv preprint arXiv:1412.6980}, 2014.

\bibitem[Kingma \& Welling(2013)Kingma and Welling]{Welling2014}
Diederik~P Kingma and Max Welling.
\newblock Auto-encoding variational bayes.
\newblock \emph{arXiv preprint arXiv:1312.6114}, 2013.

\bibitem[Kingma et~al.(2014)Kingma, Mohamed, Rezende, and Welling]{Kingma2014}
Diederik~P Kingma, Shakir Mohamed, Danilo~Jimenez Rezende, and Max Welling.
\newblock Semi-supervised learning with deep generative models.
\newblock In \emph{Advances in Neural Information Processing Systems}, pp.\
  3581--3589, 2014.

\bibitem[Kulkarni et~al.(2015)Kulkarni, Whitney, Kohli, and
  Tenenbaum]{kulkarni2015deep}
Tejas~D Kulkarni, William~F Whitney, Pushmeet Kohli, and Josh Tenenbaum.
\newblock Deep convolutional inverse graphics network.
\newblock In \emph{Advances in Neural Information Processing Systems}, pp.\
  2539--2547, 2015.

\bibitem[Larsen et~al.(2015)Larsen, S{\o}nderby, and Winther]{Larsen2015}
Anders Boesen~Lindbo Larsen, S{\o}ren~Kaae S{\o}nderby, and Ole Winther.
\newblock Autoencoding beyond pixels using a learned similarity metric.
\newblock \emph{arXiv preprint arXiv:1512.09300}, 2015.

\bibitem[Liu et~al.(2015)Liu, Luo, Wang, and Tang]{Eyebrows2015}
Ziwei Liu, Ping Luo, Xiaogang Wang, and Xiaoou Tang.
\newblock Deep learning face attributes in the wild.
\newblock In \emph{Proceedings of the IEEE International Conference on Computer
  Vision}, pp.\  3730--3738, 2015.

\bibitem[Louizos et~al.(2015)Louizos, Swersky, Li, Welling, and
  Zemel]{Louizos2015}
Christos Louizos, Kevin Swersky, Yujia Li, Max Welling, and Richard Zemel.
\newblock The variational fair auto encoder.
\newblock \emph{arXiv preprint arXiv:1511.00830}, 2015.

\bibitem[Mansimov et~al.(2015)Mansimov, Parisotto, Ba, and
  Salakhutdinov]{mansimov2015generating}
Elman Mansimov, Emilio Parisotto, Jimmy~Lei Ba, and Ruslan Salakhutdinov.
\newblock Generating images from captions with attention.
\newblock \emph{arXiv preprint arXiv:1511.02793}, 2015.

\bibitem[Ngiam et~al.(2011)Ngiam, Khosla, Kim, Nam, Lee, and Ng]{Ngiam2011a}
Jiquan Ngiam, Aditya Khosla, Mingyu Kim, Juhan Nam, Honglak Lee, and Andrew~Y
  Ng.
\newblock Multimodal deep learning.
\newblock In \emph{Proceedings of the 28th international conference on machine
  learning (ICML-11)}, pp.\  689--696, 2011.

\bibitem[Pandey \& Dukkipati(2016)Pandey and Dukkipati]{Pandey2016}
Gaurav Pandey and Ambedkar Dukkipati.
\newblock Variational methods for conditional multimodal learning: Generating
  human faces from attributes.
\newblock \emph{arXiv preprint arXiv:1603.01801}, 2016.

\bibitem[Rezende et~al.(2014)Rezende, Mohamed, and Wierstra]{Rezende2014}
Danilo~Jimenez Rezende, Shakir Mohamed, and Daan Wierstra.
\newblock Stochastic backpropagation and approximate inference in deep
  generative models.
\newblock \emph{arXiv preprint arXiv:1401.4082}, 2014.

\bibitem[Salakhutdinov \& Hinton(2009)Salakhutdinov and
  Hinton]{salakhutdinov2009deep}
Ruslan Salakhutdinov and Geoffrey~E Hinton.
\newblock Deep boltzmann machines.
\newblock In \emph{AISTATS}, volume~1, pp.\ ~3, 2009.

\bibitem[Sohn et~al.(2014)Sohn, Shang, and Lee]{Sohn2014}
Kihyuk Sohn, Wenling Shang, and Honglak Lee.
\newblock Improved multimodal deep learning with variation of information.
\newblock In \emph{Advances in Neural Information Processing Systems}, pp.\
  2141--2149, 2014.

\bibitem[Sohn et~al.(2015)Sohn, Lee, and Yan]{Sohn2015}
Kihyuk Sohn, Honglak Lee, and Xinchen Yan.
\newblock Learning structured output representation using deep conditional
  generative models.
\newblock In \emph{Advances in Neural Information Processing Systems}, pp.\
  3483--3491, 2015.

\bibitem[S{\o}nderby et~al.(2016)S{\o}nderby, Raiko, Maal{\o}e, S{\o}nderby,
  and Winther]{Sonderby2016a}
Casper~Kaae S{\o}nderby, Tapani Raiko, Lars Maal{\o}e, S{\o}ren~Kaae
  S{\o}nderby, and Ole Winther.
\newblock Ladder variational autoencoders.
\newblock \emph{arXiv preprint arXiv:1602.02282}, 2016.

\bibitem[Srivastava \& Salakhutdinov(2012)Srivastava and
  Salakhutdinov]{Srivastava2012}
Nitish Srivastava and Ruslan~R Salakhutdinov.
\newblock Multimodal learning with deep boltzmann machines.
\newblock In \emph{Advances in neural information processing systems}, pp.\
  2222--2230, 2012.

\bibitem[Team et~al.(2016)Team, Al-Rfou, Alain, Almahairi, Angermueller,
  Bahdanau, Ballas, Bastien, Bayer, Belikov,
  et~al.]{TheTheanoDevelopmentTeam2016}
The Theano~Development Team, Rami Al-Rfou, Guillaume Alain, Amjad Almahairi,
  Christof Angermueller, Dzmitry Bahdanau, Nicolas Ballas, Fr{\'e}d{\'e}ric
  Bastien, Justin Bayer, Anatoly Belikov, et~al.
\newblock Theano: A python framework for fast computation of mathematical
  expressions.
\newblock \emph{arXiv preprint arXiv:1605.02688}, 2016.

\bibitem[Yan et~al.(2015)Yan, Yang, Sohn, and Lee]{yan2015attribute2image}
Xinchen Yan, Jimei Yang, Kihyuk Sohn, and Honglak Lee.
\newblock Attribute2image: Conditional image generation from visual attributes.
\newblock \emph{arXiv preprint arXiv:1512.00570}, 2015.

\end{thebibliography}
\bibliographystyle{iclr2017_conference}

\appendix
\section{Relation between the objective of JMVAE-kl and the variation of information}
\label{sec:derivation}
The variation of information (VI) can be expressed as $-E_{p_\mathcal{D}(\mathbf{x},\mathbf{w})}[\log p(\mathbf{x}|\mathbf{w})+\log p(\mathbf{w}|\mathbf{x})]$, where $p_\mathcal{D}$ is the data distribution. In this equation, we specifically examine the sum of two negative log-likelihoods and do not consider the expectation in this derivation. We can calculate the lower bounds of these log-likelihoods as follows:
\begin{eqnarray}
	\log p(\mathbf{x}|\mathbf{w})+\log p(\mathbf{w}|\mathbf{x}) &\geq & E_{q(\mathbf{z}|\mathbf{x},\mathbf{w})}[\log \frac{p(\mathbf{x}|\mathbf{z})p(\mathbf{z}|\mathbf{w})}{q(\mathbf{z}|\mathbf{x},\mathbf{w})}] + E_{q(\mathbf{z}|\mathbf{x},\mathbf{w})}[\log \frac{p(\mathbf{w}|\mathbf{z})p(\mathbf{z}|\mathbf{x})}{q(\mathbf{z}|\mathbf{x},\mathbf{w})}]\nonumber \\
	&=& E_{q(\mathbf{z}|\mathbf{x},\mathbf{w})}[\log p(\mathbf{x}|\mathbf{z})] + E_{q(\mathbf{z}|\mathbf{x},\mathbf{w})}[\log p(\mathbf{w}|\mathbf{z})]\nonumber \\
	& & -D_{KL}(q(\mathbf{z}|\mathbf{x},\mathbf{w})||q(\mathbf{z}|\mathbf{x})) - D_{KL}(q(\mathbf{z}|\mathbf{x},\mathbf{w})||q(\mathbf{z}|\mathbf{w}))\nonumber \\
	&=& \mathcal{L}_{JM}(\mathbf{x},\mathbf{w}) - [D_{KL}(q(\mathbf{z}|\mathbf{x},\mathbf{w})||q(\mathbf{z}|\mathbf{x})) + D_{KL}(q(\mathbf{z}|\mathbf{x},\mathbf{w})||q(\mathbf{z}|\mathbf{w}))]\nonumber \\
	& & +D_{KL}(q(\mathbf{z}|\mathbf{x},\mathbf{w})||q(\mathbf{z}))\nonumber \\
	&=& \mathcal{L}_{JM_{kl}(1)}(\mathbf{x},\mathbf{w}) + D_{KL}(q(\mathbf{z}|\mathbf{x},\mathbf{w})||q(\mathbf{z})) \geq \mathcal{L}_{JM_{kl}(1)}(\mathbf{x},\mathbf{w}),
\end{eqnarray}
where $\mathcal{L}_{JM_{kl}(1)}$ is Equation \ref{eq:new_lowerbound} with $\alpha=1$. 
Therefore, maximizing Equation \ref{eq:new_lowerbound} is regarded as minimizing the VI on variational inference, i.e., maximizing the lower bounds of negative VI.

\section{Test lower bounds}
\label{sec:app1}
Two lower bounds used to estimate test marginal log-likelihood $p(\mathbf{x})$ of the JMVAE are as follows:
\begin{eqnarray}
  \mathcal{L}_{single}(\mathbf{x}) = E_{q_{\phi_x}(\mathbf{z}|\mathbf{x})}[\log \frac{p_{\theta_x}(\mathbf{x}|\mathbf{z})p(\mathbf{z})}{q_{\phi_x}(\mathbf{z}|\mathbf{x})}],
  \label{eq:JMVAE_pseudo_marginal}
\end{eqnarray}

\begin{eqnarray}
  \mathcal{L}_{multiple}(\mathbf{x}) = E_{q_{\phi}(\mathbf{z}|\mathbf{x},\mathbf{w})}[\log \frac{p_{\theta_x}(\mathbf{x}|\mathbf{z})p(\mathbf{z})}{q_{\phi}(\mathbf{z}|\mathbf{x},\mathbf{w})}].
  \label{eq:JMVAE_original_marginal}
\end{eqnarray}

We can also estimate test conditional log-likelihood $p(\mathbf{x}|\mathbf{w})$ from these two lower bounds as
\begin{eqnarray}
  \mathcal{L}_{single}(\mathbf{x}|\mathbf{w}) = E_{q_{\phi_w}(\mathbf{z}|\mathbf{w})}[\log \frac{p(\mathbf{x},\mathbf{z}|\mathbf{w})}{q_{\phi_w}(\mathbf{z}|\mathbf{w})}] = E_{q_{\phi_w}(\mathbf{z}|\mathbf{w})}[\log \frac{p_{\theta_x}(\mathbf{x}|\mathbf{z})p_{\theta_w}(\mathbf{w}|\mathbf{z})p(\mathbf{z})}{q_{\phi_w}(\mathbf{z}|\mathbf{w})}]-\log p(\mathbf{w}),
  \label{eq:JMVAE_pseudo_marginal}
\end{eqnarray}

\begin{eqnarray}
  \mathcal{L}_{multiple}(\mathbf{x}|\mathbf{w}) = E_{q_{\phi}(\mathbf{z}|\mathbf{x},\mathbf{w})}[\log \frac{p_{\theta_x}(\mathbf{x}|\mathbf{z})p_{\theta_w}(\mathbf{w}|\mathbf{z})p(\mathbf{z})}{q_{\phi}(\mathbf{z}|\mathbf{x},\mathbf{w})}]-\log p(\mathbf{w}),
  \label{eq:JMVAE_original_marginal}
\end{eqnarray}
where $\log p(\mathbf{w}) = \log E_{p(\mathbf{z})}[p_{\theta_w}(\mathbf{w}|\mathbf{z})] = \log \frac{1}{N_w}\sum^{N_w}_i p_{\theta_w}(\mathbf{w}|\mathbf{z}^{(i)})$ and $\mathbf{z}^{(i)}\sim p(\mathbf{z})$. In this paper, we set $N_w=5,000$ on MNIST and $N_w=10$ on CelebA.

We can obtain a tighter bound on the log-likelihood by $k$-fold importance weighted sampling. For example, we obtain an importance weighted bound on $\log p(\mathbf{x})$ from Equation \ref{eq:JMVAE_pseudo_marginal} as follows:
\begin{eqnarray}
  \log p(\mathbf{x}) \geq E_{{\mathbf{z}_1},...,{\mathbf{z}_k}\sim q_{\phi_x}(\mathbf{z}|\mathbf{x})}[\log\frac{1}{k}\sum_{i=1}^k \frac{p_{\theta_x}(\mathbf{x}|\mathbf{z})p(\mathbf{z})}{q_{\phi_x}(\mathbf{z}|\mathbf{x})}]=\mathcal{L}^k_{single}(\mathbf{x}).
  \label{eq:JMVAE_pseudo_marginal}
\end{eqnarray}

Strictly speaking, these two lower bounds are not equal. However, if the number of importance samples is extremely large, the difference of these two lower bounds converges to $0$.
\begin{proof}
Let the multiple and single $k$-hold importance weighted lower bounds as $\mathcal{L}^k_{single}$ and $\mathcal{L}^k_{single}$. From the theorem of the importance weighted bound, both $\mathcal{L}^k_{single}$ and $\mathcal{L}^k_{multiple}$ converge to $\log p(\mathbf{x})$ as $k\rightarrow \infty$.

Therefore,
\begin{eqnarray*}
{\lim}_{k\rightarrow \infty}|\mathcal{L}_{multiple}^k-\mathcal{L}_{single}^k|\leq |{\lim}_{k\rightarrow \infty}\mathcal{L}_{multiple}^k-{\lim}_{k\rightarrow \infty}\mathcal{L}_{single}^k|=0 \qedhere
\end{eqnarray*}
\end{proof}

\section{Reconstructed images}
\label{sec:app2}

\begin{figure}[t]
 \begin{center}
  \includegraphics[scale=0.65]{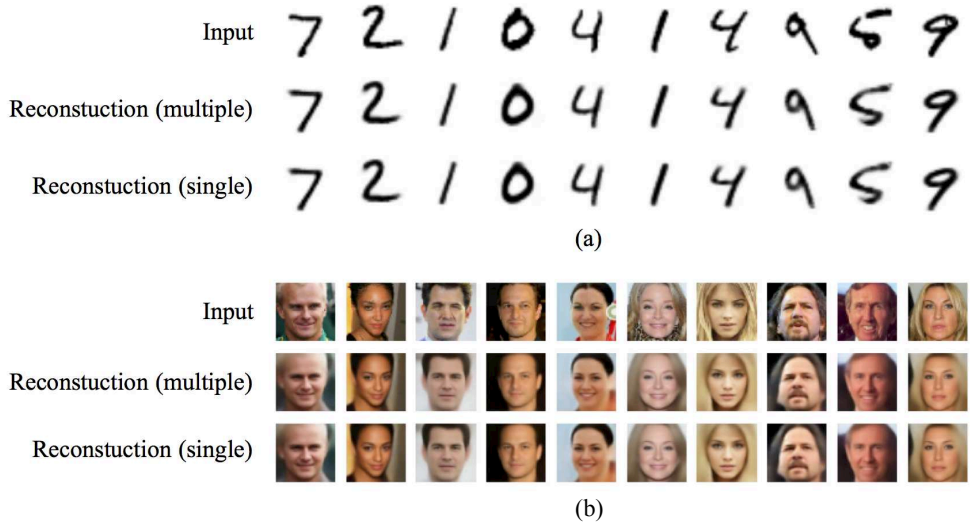}
 \end{center}
 \caption{Comparison of the original images and reconstructed images by the JMVAE ($\alpha=0.1$). We used (a) MNIST and (b) CelebA datasets.}
 \label{fig:px_reconst}
\end{figure}

Figure \ref{fig:px_reconst} presents a comparison of the original image and reconstructed image by the JMVAE on both MNIST and CelebA datasets. It is apparent that the JMVAE can reconstruct the original image properly with either a multiple or single encoder.

\section{Test joint log-likelihood on MNIST}

\begin{table}[tb]
   \caption{Evaluation of test log-likelihood. All models are trained on the MNIST dataset: {\sl left}, marginal log-likelihood; {\sl right}, conditional log-likelihood.}
  \begin{center}
  \small
  \begin{tabular}{lc}\hline
     & $\leq\log p(\mathbf{x},\mathbf{w})$ \\\hline
     JMVAE-zero & -86.96 \\
     JMVAE-kl, $\alpha=0.01$ & -86.94 \\
     JMVAE-kl, $\alpha=0.1$ & -86.93 \\
     JMVAE-kl, $\alpha=1$ & -89.28 \\\hline
  \end{tabular}
  \label{tab:mnist_marginal_log-likelihood}
  \end{center}
\end{table}

Table \ref{tab:mnist_marginal_log-likelihood} shows the joint log-likelihood of the JMVAE on MNIST dataset by both JMVAE-zero and JMVAE-kl. It is apparent that the log-likelihood test on both approaches is almost identical (strictly, JMVAE-zero is slightly lower). The test log-likelihood on JMVAE-kl becomes much lower if $\alpha$ is large.

\section{Image generation from conditional distribution on MNIST}
\label{sec:mnist_cond}
\begin{figure}[tb]
\begin{center}
  \includegraphics[scale=0.55]{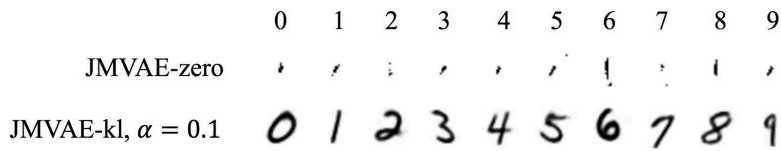}
 \end{center}
 \caption{Image generation from conditional distribution $p(\mathbf{x}|\mathbf{w})$. We used a single encoder $p(\mathbf{z}|\mathbf{w})$ for both generations.}
 \label{fig:px_w_reconst_mnist}
\end{figure}

Figure \ref{fig:px_w_reconst_mnist} presents generation samples of $\mathbf{x}$ conditioned on single input $\mathbf{w}$. It is apparent that the JMVAE with JMVAE-kl generates conditioned digit images properly, although that with JMVAE-zero cannot generate them. As results showed, we also confirmed qualitatively that JMVAE-kl can model $q_{\phi_x}(\mathbf{z}|\mathbf{x})$ properly compared to JMVAE-zero.

\end{document}